\documentclass{article}

\usepackage{lipsum} 

\usepackage{graphicx}

\usepackage{caption}
\usepackage{xcolor}%

\usepackage{adjustbox}
\usepackage{float}
\usepackage{multirow}
\usepackage{subcaption}

\usepackage{hyperref}

\usepackage{booktabs}

\usepackage{todonotes}

\usepackage{amsmath,amssymb,amsfonts}%
\usepackage{amsthm}%

\usepackage{algorithmicx}%
\usepackage{algpseudocode}%
\usepackage{listings}%

\usepackage{svg}
\usepackage[section]{placeins}

\begin{document}

\title{A Simple Architecture for Enterprise Large Language Model Applications based on Role based security and Clearance Levels using Retrieval-Augmented Generation or Mixture of Experts}

\author{Atilla Özgür$^1$\thanks{Atilla Özgür is partially supported by Grant DFG UY-56/5-1} \and Yılmaz Uygun $^1$ }

\date{
	$^1$School of Business, Social and Decision Sciences, Constructor University Bremen, Bremen, Germany  \\ \texttt{\{aoezguer, yuygun\}@constructor.university}
}

\maketitle

\begin{abstract}

This study proposes a simple architecture for Enterprise application for Large Language Models (LLMs) for role based security and NATO clearance levels.
Our proposal aims to address the limitations of current LLMs in handling security and information access.
The proposed architecture could be used while utilizing Retrieval-Augmented Generation (RAG) and fine tuning of Mixture of experts models (MoE).
It could be used only with RAG, or only with MoE or with both of them.
Using roles and security clearance level of the user, documents in RAG and experts in MoE are filtered.
This way information leakage is prevented.


\end{abstract}

\section{Introduction}
\label{section-introduction}

According to World Economic Forum \cite{Forum2024How}, Venture Capital investments in the area of artificial intelligence are about \$290 billion in the last 5 years.
With the introduction of ChatGPT by OpenAI \cite{OpenAI2022ChatGPT}, interest in the large language models (LLMs) has exploded.

Even though ChatGPT provides an application programming interface (API) to the developers, it is a closed model.
Developers could only utilize provided API. 
To answer ChatGPT from OpenAI, large language model called LLaMA \cite{Touvron2023LLaMA} is introduced by Meta (Facebook).
With the advent of LLaMA, the era of open source Large Language Models started.
Open source developers trained their own models in different domains using LLaMA and published these models.
Due to success of this, other companies followed and published their own models in open source fashion.
For example, Microsoft published Phi \cite{Abdin2024Phi} and Google published gemini \cite{GemmaTeam2024Gemma} open source LLMs.
A more complete list could be found in \cite{Yan2024Open}.

Additionally, a lot of different client libraries for these open source models are introduced.
With these client libraries, it is easier than ever to produce applications which utilizes LLMs in their portfolio.
Security of the these applications are very important.
But in academia, instead of general security, most of the time security for LLM itself are the focus.
Security of LLMs are reviewed in a lot of different articles \cite{Yao2024survey,Chowdhury2024Breaking,Zhang2024When,Das2024Security,Liu2024Review,Hassanin2024Comprehensive,Ferrag2024Generative,Goto2024Evaluating,Wang2024Unique,Luo2024Privacy,Wu2024New,Pankajakshan2024Mapping,Derner2023Security,Iqbal2023LLM,Lin2023AgentSims}.
But these articles are mostly about the attack on LLMs itself.
Even articles about privacy is mostly concerned from general usage of LLMs.
According to the best of our knowledge, how to secure LLMs in enterprises from the view point of security clearance levels and role based security is not investigated.

A recent lawsuit by New York Times \cite{Grynbaum2023Times} shows that is is possible to reproduce exact trained documents using appropriate prompts.
Normally, this is only a copyright and privacy problem but in the context of military or enterprise applications, such a problem becomes a big security problem.
Think about the following example: A custom LLM application developed for NATO military documents.
If an user of this NATO LLM application could reproduce a Secret NATO document when his security clearance is only for Confidential documents, this reproduction of document would be a very big security problem.

This study proposes a simple role based security architecture for custom LLM applications.
This approach could be used while utilizing Retrieval-Augmented Generation (RAG) and fine tuning of Mixture of experts models (MoE).
It could be used only with RAG, or only with MoE or with both of them.
While using RAG, only documents which user's roles have access to will be returned from RAG.
While using MoE, only experts which user's roles have access will be consulted.
This way, information leakage is prevented and security of the application will be ensured.

\section{Background Information}
\label{section-background-information}

To be able to understand proposed model more easily, some background information is needed.
In this section, this background information about NATO clearance levels, role based security, introduction to large language models, Retrieval-Augmented Generation (RAG) and Mixture of experts are given.

\subsection{NATO Clearance Levels}
\label{section-clearance-levels}

The North Atlantic Treaty Organization (NATO) is a military alliance.
NATO consists of 32 member states and is established after World War II.
NATO like many international organizations deals with sensitive information. 
This information can range from battle plans to diplomatic communications and intelligence reports.
Leaking this information could have serious consequences \cite{Comittee1958Short}.

NATO has following four security classifications:

\begin{enumerate}
	\item Cosmic top secret
	\item Secret
	\item Confidential
	\item Restricted
\end{enumerate}

Even though it is not counted among the classifications, not classified category is also used to show that a document or an information could be shared outside  NATO but their rights belong to NATO \cite{Roberts2002Nato}.

Clearance levels or security classifications are used for following purposes: Protect Classified Information, Minimize Risks and Ensure Trust Between Allies
Classified information is categorized according to its sensitivity.
For example size information of F-35 fighter jet could be not classified while radar sensitive painting information of F-35 could be Cosmic top secret.
In short, NATO clearance levels are a security categorization designed to prevent sensitive information.

\subsection{Role-Based Access Control (RBAC)}
\label{section-role-based-access-control}

Role-Based Access Control (RBAC) is a security approach that manages access to resources within a system.
Idea of RBAC has been around since the start of multi user computers \cite{Sandhu1998Role}.
RBAC resolves around permissions, roles and user.

\begin{figure}[!htbp]
	\includegraphics[width=\textwidth]{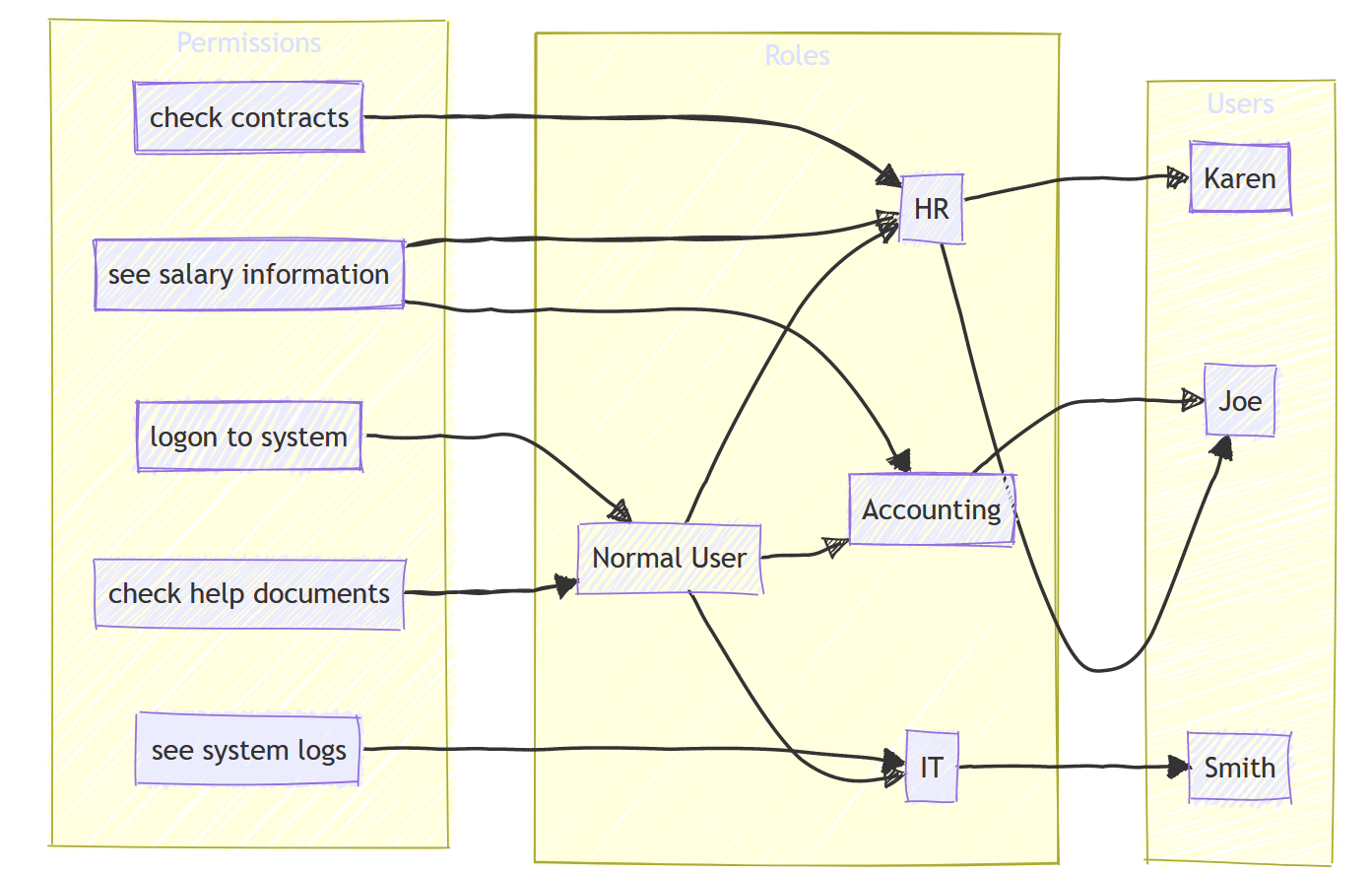}
	\caption{Example Role Based Access}
	\label{figure-role-based-example1}
\end{figure}

Permissions control access to resources or abilities of users. 
An example for permission could be log on to system.
Roles are used as containers for permissions.
For example a human resources (HR) role could be used to contain permission related to personnel management tasks.
In most systems, a role could also contain other roles.
Figure~\ref{figure-role-based-example1} shows that Normal User role belongs to other roles.
Similar to Roles, Users also could belong to more than one role.

When security clearance levels are used, most of the time, these clearance levels names are extended together with normal roles.
For example, Operator role will be extended as Operator not classified, Operator restricted and so on.

\subsection{Introduction to large language models}
\label{section-intro-large-language-models}

Large Language Models are supervised learning models.
That is they are trained with using known input and outputs.
LLMs are trained using a lot of text from internet to repeatedly predict next words using previous words.
See Table~\ref{table-llm-text-generation-example} for an example sentence: My favorite food is a Döner with spicy sauce.

\begin{table}[!htbp]
\caption{LLM Text Generation Example}
\label{table-llm-text-generation-example}
\begin{tabular}{@{}p{0.8\textwidth}p{0.2\textwidth}@{}}

\toprule

Input A                                & Output \\ 

\midrule

My favorite food is a                  & Döner  \\
My favorite food is a Döner            & with   \\
My favorite food is a Döner with       & spicy  \\
My favorite food is a Döner with spicy & sauce  \\ 

\bottomrule

\end{tabular}
\end{table}

\subsubsection{Training of large language models}

For training LLMs, terabytes of text from internet are used.
Basic workflow for training LLM are something like below:

\begin{enumerate}
	\item Download ~10+TB of text
	\item Get a cluster of 6k+ GPUs
	\item Train your neural network, pay ~\$2M, wait for ~12 days
	\item Obtain \textbf{base model}
\end{enumerate}

This training could be seen as something like compressing the training dataset.
Since LLMs could only produce values that are in their training data, for specialized tasks fine tuning of LLMs should be done.
Fine tuning should be done with quality data specially prepared for the task.
Basic workflow for fine tuning LLM is something like below:

\begin{enumerate}
	\item create curated dataset of quality instructions
	\item fine tune base model on this data, wait ~1 day
	\item Obtain \textbf{assistant model}
	\item Test your model if necessary go to step 1
	\item Deploy on your servers
	\item Monitor, go to step 1
\end{enumerate}

Since fine tuning uses less data, it costs less and could be repeated more.
When using Mixture of experts, first fine tuning would be slower since we will need to fine tune multiple expert LLMs.
But subsequent fine tuning for experts will be faster since whenever new documents come only the relevant experts will be re-fine tuned.

\subsection{Retrieval-Augmented Generation (RAG)}
\label{section-retrieval-augmented-generation}

As we have explained in section~\ref{section-intro-large-language-models}, LLMs are trained with massive amounts of data.
As all machine learning models, LLMs are also depended on statistical patterns in their training data.
For example, an LLM model, which is trained in 2023, will not be able to answer questions about Euro 2024 (European Football Championship).

Retrieval-Augmented Generation (RAG) introduced by Facebook researchers\cite{Lewis2020Retrieval} address these limitation by connecting LLMs to update data sources. 
These sources could be news articles, company internal knowledge base or databases like wikipedia.

\begin{figure}[!htbp]
	\includegraphics[width=\textwidth]{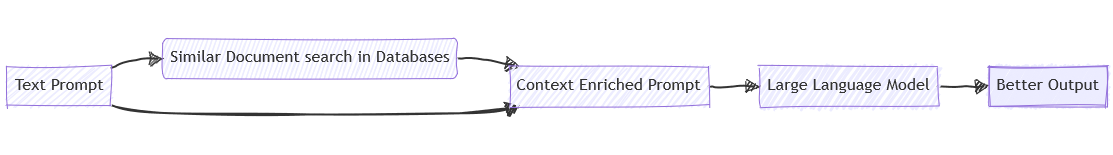}
	\caption{Retrieval-Augmented Generation (RAG) workflow}
	\label{figure-RAG-workflow}
\end{figure}

RAG workflow could be seen in Figure~\ref{figure-RAG-workflow}

When a new prompt comes to LLM system, similar documents to this new prompt are searched in databases.
Most of the time a vector database is used for fast response times.
Then, LLM uses this context enriched prompt to give answers.

RAG makes LLM outputs more reliable using factual databases.
Like previous example of Euro 2024, RAG enables LLMs to use latest information if their training data is older.
RAG could also be adapted to specific domains using relevant databases.

In our proposed role-based access control architecture, every document in RAG databases will have allowed Roles information.
When similar documents are searched for RAG prompts, only documents which asking user have access to will be searched.
This way, information leakage will be prevented.

\subsection{Mixture of experts}
\label{section-mixture-of-experts}

Jacobs et al \cite{Jacobs1991Adaptive} introduced mixture-of-experts (MoE) model in 1991.
The MoE models utilizes individual neural networks as experts instead of a single neural network.
The idea of mixture of experts are very similar to ensemble learning \cite{Polikar2009Ensemble}.
Ensemble learning combines multiple classifiers in different fashion.
According to Polikar \cite{Polikar2009Ensemble}, among the first examples of Ensemble Learning \cite{Dasarathy1979composite} was in 1979 by Dasarathy and Sheela.
Mixture of experts name is more popular in deep learning literature while ensemble learning is more popular in general machine learning literature.

In mixture of experts, different neural networks to solve the problem are called experts.
Mixture of expert model uses a router or coordinator neural network to decide which experts to utilize \cite{Masoudnia2012Mixture}.
Router assigns a "gating score" to each network to indicate how relevant the input to the expert is.
A softmax function is used to transform the gating scores to a probability distribution.
When doing the predictions only experts with highest gating scores are activated.
This router idea is very similar to voting in the ensemble classifiers.
Voting classifiers are used in a lot of different domains like intrusion detection \cite{Zeghida2023Securing,Ali2023Effective,Saheed2024voting,Oezguer2018Sparsity,Oezguer2018Feature}

In our role based access control architecture, gating scores of  some experts will be zero according to the roles of the caller.
For example, a normal user is calling the large language model with a human resource prompt. 
Only the experts which normal user has access to will be activated.
Other experts like human resources expert will not be called at all.

\section{Proposed Architecture based on Role
based security and Clearance Levels}
\label{section-methods}

\subsection{User, Role, Security Clearance to Documents Mapping}
\label{section-role-document-mapping}

\begin{figure}[!htbp]
	\includegraphics[width=\textwidth]{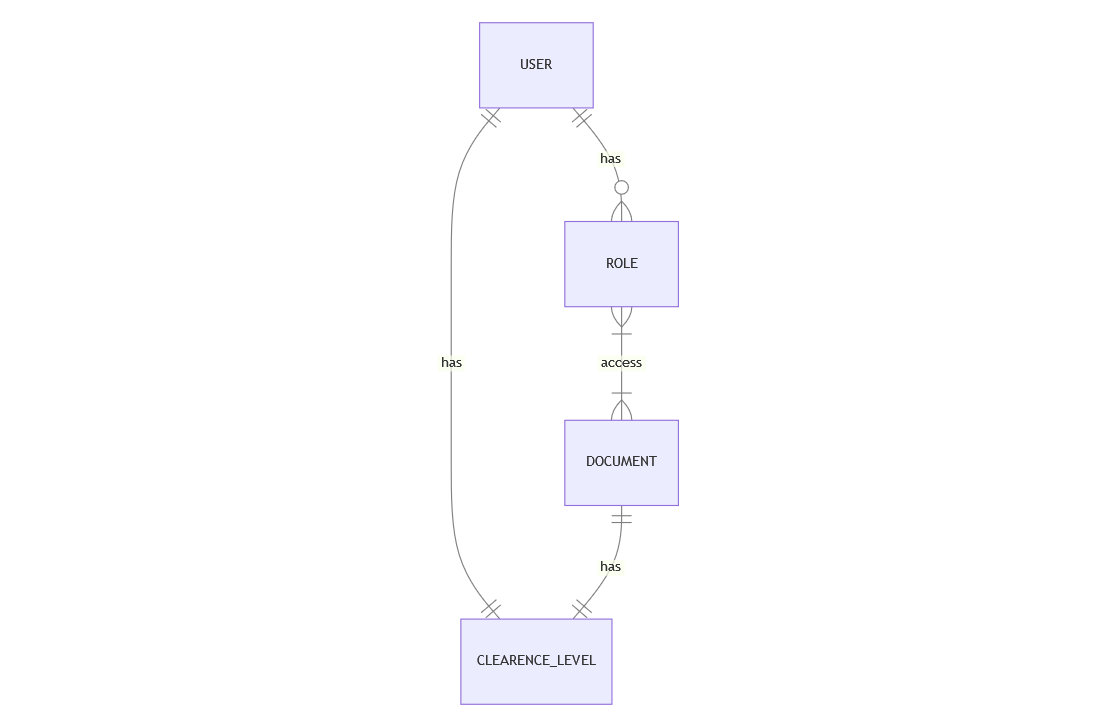}
	\caption{Entity Relationship Diagram for User to Role and Clearance Level to Document Mapping}
	\label{figure-role-document-db-diagram}
\end{figure}

Most enterprises map their users to one or more roles.
This mapping could be stored in directory services like active directory (Microsoft) or Enterprise Resource planning databases.
For our LLM application, this user to role and user to security clearance levels mappings should be accessible from its programming interface.
Additionally, role to documents and security clearance levels to documents should also be accessible.
This mapping information could be stored in LLM application's own database or a web service could be provided to the application.
Basic mapping could be seen in Figure~\ref{figure-role-document-db-diagram}.

Every User has more than one role, every User has exactly one security clearance.
Every Document has exactly one security clearance.
Every Role has access to zero or more documents.
It is implied that if a user does not have access to necessary clearance level, he will not be able to access the document.

\subsection{Training LLM using Roles and Security Clearance Levels}
\label{section-training-LLM}

Proposed architecture presumes using local open source LLMs due to security requirements.
But if strict local security is not required then commercial LLMs like OpenAI ChatGPT with Retrieval-Augmented Generation (RAG) could be used.
When commercial LLMs are used, training step may not be necessary according to requirements or only not confidential documents could be used for training.

For local open source LLMs, Mixture of Experts model should be used.
For this case, it is advised to train multiple experts for every role multiplied by five, that is four clearance levels plus not classified.
For example, let's assume that the LLM application has four roles: HR, Accounting, Normal User, IT  as shown in Figure~\ref{figure-role-based-example1}.
Then, we should train $4*5=20$ experts.
This experts will be named like HR Not Classified, HR Restricted, HR Confidential, HR Secret, HR Cosmic top secret.
See full names in Table~\ref{table-roles-vs-clearance-levels}.

\begin{table}[!htbp]
\caption{Roles and Clearance Levels for Experts}
\label{table-roles-vs-clearance-levels}
\begin{tabular}{@{}p{0.2\textwidth}p{0.2\textwidth}p{0.2\textwidth}p{0.2\textwidth}p{0.2\textwidth}@{}}

\toprule

                  & HR                & Accounting        & Normal User       & IT                \\ 

\midrule
Not classified    & HR Not classified    & Accounting Not classified    & Normal User Not classified    & IT Not classified    \\
Restricted        & HR Restricted        & Accounting Restricted        & Restricted        & IT Restricted        \\
Confidential      & HR Confidential      & Accounting Confidential      & Normal User Confidential      & IT Confidential      \\
Secret            & HR Secret            & Accounting Secret            & Normal User Secret            & IT Secret            \\
Cosmic top secret & HR Cosmic top secret & Accounting Cosmic top secret & Normal User Cosmic top secret & IT Cosmic top secret \\ 
\bottomrule
\end{tabular}
\end{table}

\subsection{Inference using Retrieval-Augmented Generation (RAG)}
\label{section-inference-using-RAG}

While doing the inference using Retrieval-Augmented Generation (RAG), role information and security clearance levels should be used.
Most of the vector databases allow use of filters.
Using filters, only documents a user has access to should be returned.

\begin{figure}[!htbp]
	\includegraphics[width=\textwidth]{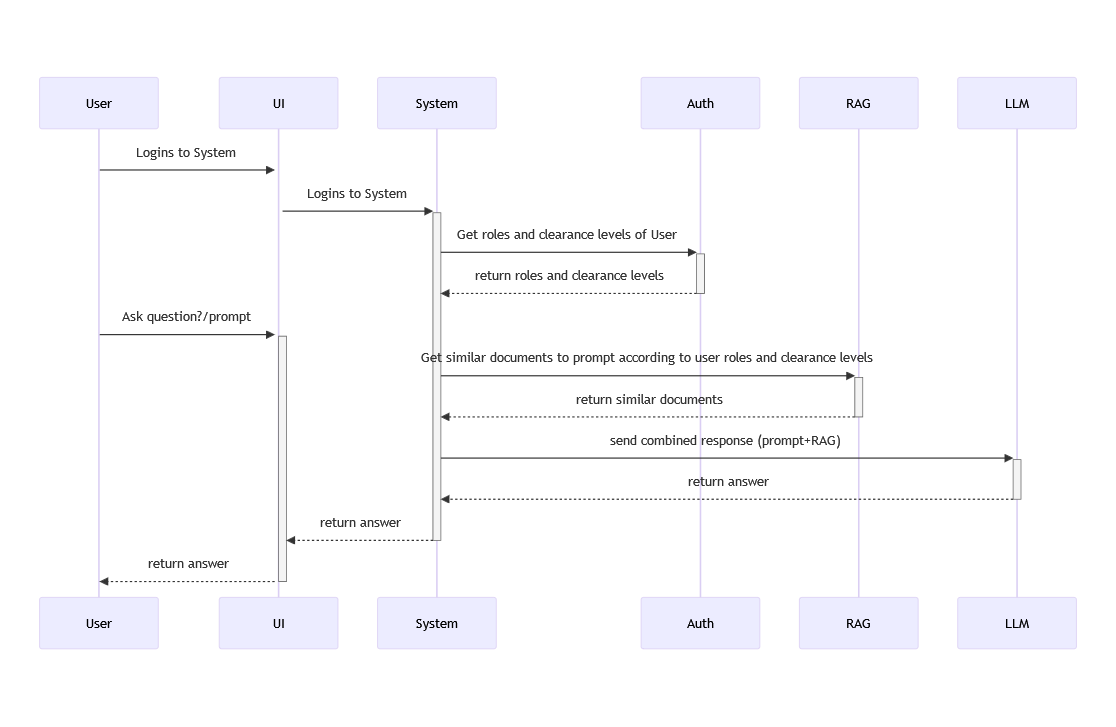}
	\caption{ Sequence Diagram for Role/Clearance level based access to LLM only using RAG}
	\label{figure-role-based-access-to-LLM-only-RAG-sequence-diagram}
\end{figure}

As could be seen in Figure~\ref{figure-role-based-access-to-LLM-only-RAG-sequence-diagram}, this architecture does not presume mixture of experts or local open source model; therefore, this role based model could also be used with commercial LLMs like ChatGPT.

\subsection{Inference using Mixture of experts models (MoE)}
\label{section-inference-using-MoE}

\begin{figure}[!htbp]
	\includegraphics[width=\textwidth]{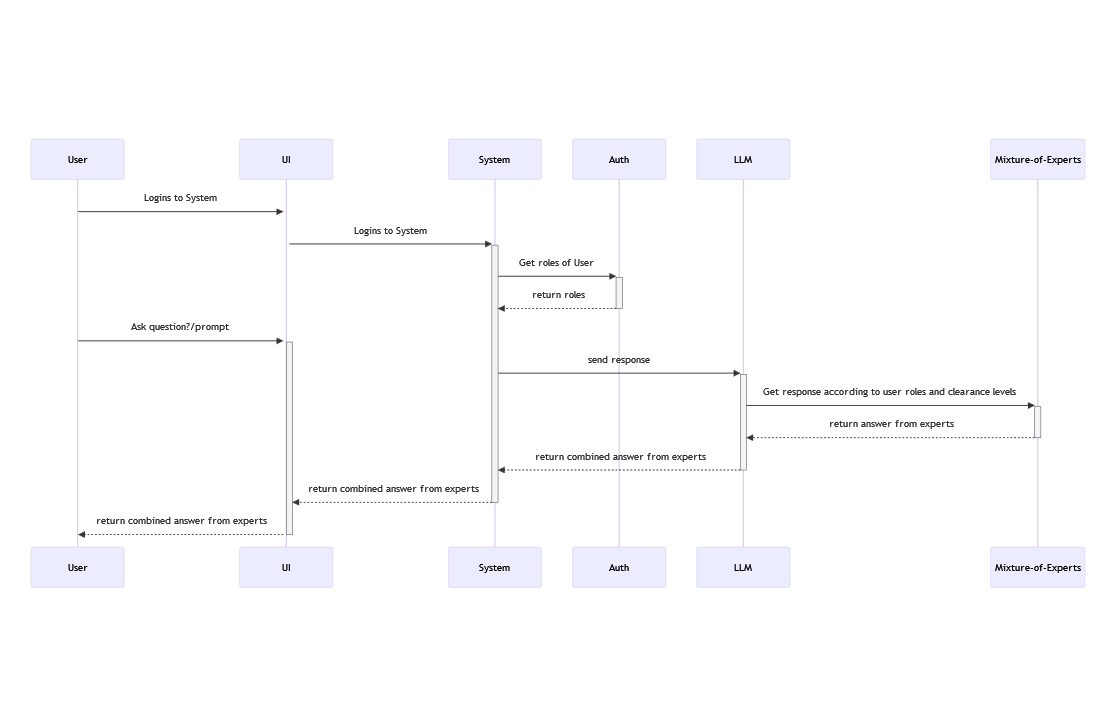}
	\caption{ Sequence Diagram for Role/Clearance level based access to LLM only using MoE}
	\label{figure-role-based-access-to-LLM-only-MoE-sequence-diagram}
\end{figure}

In this inference model, local open source LLM should be used.
Experts in MoE are trained using only documents for which role and security clearance level has an access to.
For performance reasons, filtering in the router part of MoE would be useful.
If router does not ask answers from experts for whom user has no security clearance or necessary role, the application will perform faster.
Full workflow could be seen in Figure~\ref{figure-role-based-access-to-LLM-only-MoE-sequence-diagram}.

\subsection{Inference using both RAG and MoE}
\label{section-inference-using-both-RAG-and-MoE}

\begin{figure}[!htbp]
	\includegraphics[width=\textwidth]{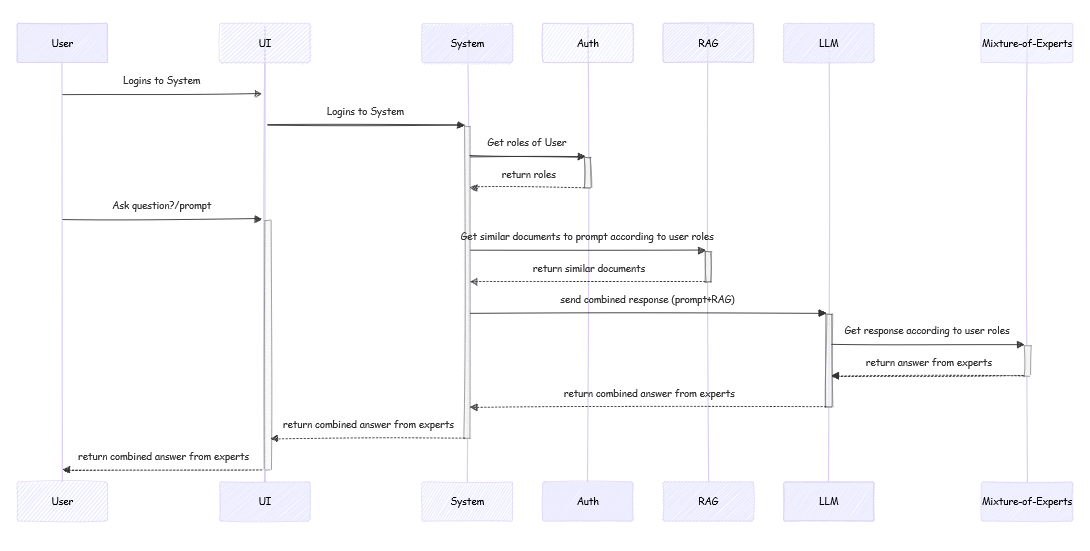}
	\caption{Role based access to LLM Sequence Diagram}
	\label{figure-role-based-access-to-LLM-sequence-diagram}
\end{figure}

Doing inference using both Retrieval-Augmented Generation (RAG) and Mixture of experts models (MoE) could be seen in Figure~\ref{figure-role-based-access-to-LLM-sequence-diagram}.
Basically, this workflow is combination of previous two workflows.

\section{Conclusion}
\label{section-conclusion}

A simple role based security architecture for Large Language Model (LLM) applications are proposed.
Background information necessary for the understanding of the architecture given in multiple sections.
Proposed approach is usable with  Retrieval-Augmented Generation (RAG) and/or Mixture of experts models (MoE).
Normally usage of local open source LLM models are assumed but RAG version is also suitable for commercial LLMs.
Sequence diagrams for workflows using RAG, MoE and RAG+MoE are given.

\end{document}